\documentclass[11pt,a4paper]{article}
\usepackage[hyperref]{emnlp2020}
\usepackage{times}
\usepackage{latexsym}
\usepackage{amsmath}
\usepackage{graphicx}
\usepackage{multirow}
\usepackage{booktabs}
\usepackage{arydshln}
\usepackage{verbatim}
\usepackage{changepage}

\aclfinalcopy 

\title{{O}nline {B}ack-{P}arsing for {A}MR-to-{T}ext {G}eneration}

\author{
	Xuefeng Bai$^{\spadesuit \heartsuit}$\hspace{0.5mm}, 
	Linfeng Song$^{\clubsuit}$\hspace{0.5mm}, 
	Yue Zhang$^{\heartsuit \diamondsuit}$\hspace{0.2mm} \\
	$^\spadesuit$ Zhejiang University, China \\
	$^\heartsuit$School of Engineering, Westlake University, China \\
	$^\clubsuit$ Tencent AI Lab, Bellevue, WA, USA \\
	$^\diamondsuit$Institute of Advanced Technology, Westlake Institute for Advanced Study, China\\	
}

\date{}

\begin{document}
\maketitle
\begin{abstract}
	AMR-to-text generation aims to recover a text containing the same meaning as an input AMR graph.
    Current research develops increasingly powerful graph encoders to better represent AMR graphs, with decoders based on standard language modeling being used to generate outputs.
    We propose a decoder that back predicts projected AMR graphs on the target sentence during text generation.
	As the result, our outputs can better preserve the input meaning than standard decoders.
	Experiments on two AMR benchmarks show the superiority of our model over the previous state-of-the-art system based on graph Transformer.
\end{abstract}

\section{Introduction}

\begin{figure}
	\centering
	\includegraphics[width=0.55\linewidth]{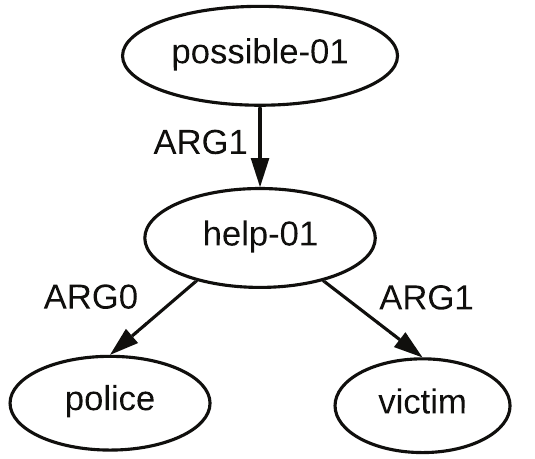}
	\caption{An example AMR graph meaning ``\emph{The police could help the victim.}''}
	\label{fig:example}
\end{figure}

Abstract meaning representation (AMR) \cite{banarescu2013abstract} is a semantic graph representation that abstracts meaning away from a sentence.
Figure \ref{fig:example} shows an AMR graph, where the nodes, such as ``\emph{possible-01}'' and ``\emph{police}'', represent concepts, and the edges, such as ``\emph{ARG0}'' and ``\emph{ARG1}'', indicate  relations between the concepts  they connect. The task of AMR-to-text generation~\cite{konstas2017neural} aims to produce fluent sentences that convey consistent meaning with input AMR graphs.
For example, taking the AMR in Figure \ref{fig:example} as input, a model can produce the sentence ``\emph{The police could help the victim}''.
AMR-to-text generation has been shown useful for many applications such as machine translation \cite{song2019semantic} and summarization \cite{liu2015toward,yasunaga2017graph,liao2018abstract,hardy2018guided}.
In addition, AMR-to-text generation can be a good test bed for general graph-to-sequence problems~\cite{belz2011first,gardent2017webnlg}.

AMR-to-text generation has attracted increasing research attention recently. Previous work has focused on developing effective encoders for representing graphs.
In particular, graph neural networks \cite{beck2018graph,song2018graph,guo-etal-2019-densely} and richer graph representations \cite{damonte2019structural,hajdik2019neural,ribeiro2019enhancing} have been shown to give better performances than RNN-based models \cite{konstas2017neural} on linearized graphs.
Subsequent work exploited graph Transformer \cite{zhu2019modeling,cai2020graph,wang2020amr}, achieving better performances by directly modeling the intercorrelations between distant node pairs with relation-aware global communication.
Despite the progress on the {\it encoder} side, the current state-of-the-art models use a rather standard {\it decoder}: it functions as a language model, where each word is generated given only the previous words.
As a result, one limitation of such decoders is that they tend to produce fluent sentences that may not retain the meaning of input AMRs.

We investigate enhancing AMR-to-text decoding by integrating online back-parsing, simultaneously predicting a {\it projected} AMR graph on the target sentence while it is being constructed. This is largely inspired by work on back-translation \cite{sennrich2016improving,tu2017neural}, which shows that back predicting the source sentence given a target translation output can be useful for strengthening neural machine translation. We perform {\it online} back parsing, where the AMR graph structure is constructed through the autoregressive sentence construction process, thereby saving the need for training a separate AMR parser. By adding online back parsing to the decoder, structural information of the source graph can intuitively be better preserved in the decoder network.

\begin{figure*}
	\centering
	\includegraphics[width=0.9\textwidth]{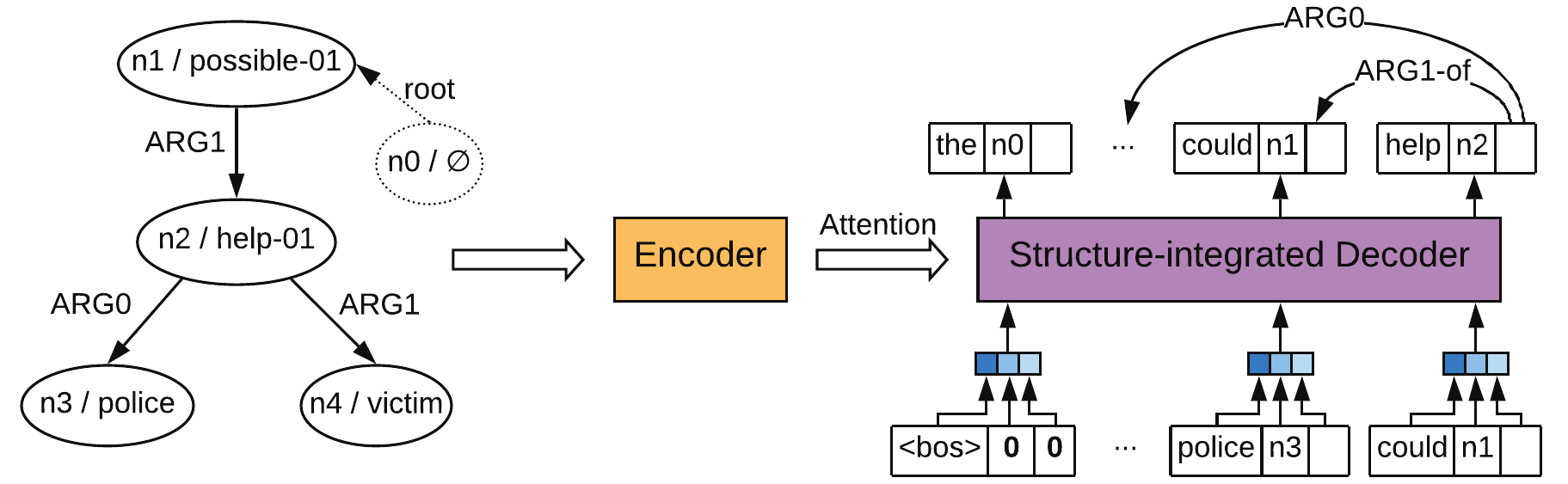}
	\caption{Overview of the proposed model.}
	\label{fig:model}
\end{figure*}

Figure \ref{fig:model} visualizes our structure-integrated decoding model when taking the AMR in Figure \ref{fig:example} as input.
In particular, at each decoding step, the model predicts the current word together with its corresponding AMR node and outgoing edges to the previously generated words.
The predicted word, AMR node and edges are then integrated as the input for the next decoding step.
In this way, the decoder can benefit from both more informative loss via multi-task training and richer features taken as decoding inputs.

Experiments on two AMR benchmark datasets (LDC2015E86 and LDC2017T10\footnote{http://amr.isi.edu/}) show that our model significantly outperforms a state-of-the-art graph Transformer baseline by 1.8 and 2.5 BLEU points, respectively, demonstrating the advantage of structure-integrated decoding for AMR-to-text  generation.
Deep analysis and human evaluation also confirms the superiority of our model.
Our code is available at \url{https://github.com/muyeby/AMR-Backparsing}.

\section{Baseline: Graph Transformer}

Formally, the AMR-to-text generation task takes an AMR graph as input, which can be denoted as a directed acyclic graph $G = (V,E)$, where $V$ denotes the set of nodes and $E$ refers to the set of labeled edges.
An edge can further be represented by a triple $\left \langle v_i, {r_k}, v_j \right \rangle $, showing that node $v_i$ and $v_j$ are connected by relation type $r_k$.
Here $k\in [1,...,R]$, and $R$ is the total number of relation types.
The goal of AMR-to-text generation is to generate a word sequence $\textbf{y} = [y_1,y_2,\dots, y_M]$, which conveys the same meaning as $G$.

We take a graph Transformer model~\cite{koncel2019text,zhu2019modeling,cai2020graph,wang2020amr} as our baseline. Previous work has proposed several variations of graph-Transformer.
We take the model of~\citet{zhu2019modeling}, which gives the state-of-the-art performance.
This approach exploits a graph Transformer encoder for AMR encoding and a standard Transformer decoder for text generation.

\subsection{Graph Transformer Encoder}
\label{sec:graphencoder}
The Graph Transformer Encoder is an extension of the standard Transformer encoder~\cite{vaswani2017attention}, which stacks $L$ encoder layers,  each having two sublayers: a self-attention layer and a position-wise feed forward layer.
Given a set of AMR nodes $[v_1, v_2, \dots, v_N]$, the $l$-th encoder layer takes the node features $[h_1^{l-1}, h_2^{l-1},\dots,h_N^{l-1}]$ from its preceding layer as input and produces a new set of features $[h_1^{l}, h_2^{l},\dots,h_N^{l}]$ as its output.
Here $h_i^{l-1},h_i^{l} \in \mathcal{R}^d$, $d$ is the feature dimension, $l \in [1,\dots,L]$, and $h_i^{0}$ represents the embedding of AMR node $v_i$, which is randomly initialized. 

The graph Transformer encoder extends the vanilla self-attention ({SAN}) mechanism by explicitly encoding the
relation $r_k$\footnote{Since the adjacency matrix is sparse, the graph Transformer encoder uses the shortest label path between two nodes to represent the relation (e.g. path (victim, police) = ``$\uparrow$ARG1 $\downarrow$ARG0'', path (police, victim) = ``$\uparrow$ARG0 $\downarrow$ARG1'').} between each AMR node pair $(v_i, v_j)$ in the graph. 
In particular, the relation-aware self-attention weights are obtained by:
\begin{equation}
\begin{split}
\alpha_{ij} &= \frac{\exp(e_{ij})}{\sum\nolimits_{n \in [1,\dots, N]} \exp{(e_{in})}}, \\
e_{ij} &= \frac{(W^Qh_{i}^{l-1})^{T}(W^{K}h_{j}^{l-1}+W^R\gamma_{k})}{\sqrt{d}},\\
\end{split}
\label{eq:relation-dotattention}
\end{equation}
where $W^Q, W^K, W^R$ are model parameters, and $\gamma_{k} \in \mathcal{R}^{d_r}$ is the embedding of relation $r_k$, which is randomly initialized and optimized during training, $d_r$ is the dimension of relation embeddings. 

With $\alpha_{ij}$, the output features are:
\begin{equation}
h^{l}_{i} = \sum\nolimits_{j \in [1,\dots,N]} \alpha_{ij}(W^{V}{h}^{l-1}_{j} + W^R\gamma_{k}),
\label{eq:relation-update}
\end{equation}
where $W^V$ is a parameter matrix.

Similar to the vanilla Transformer, a graph Transformer also uses multi-head self-attention, residual connection and layer normalization.
\subsection{Standard Transformer Decoder}
\label{sec:stddecoder}
The graph Transformer decoder is identical to the vanilla Transformer~\cite{vaswani2017attention}.
It consists of an embedding layer, multiple Transformer decoder layers and a generator layer (parameterized with a linear layer followed by softmax activation). 
Supposing that the number of decoder layers is the same as the encoder layers, denoted as $L$.
The decoder consumes the hidden states of the top-layer encoder $H^L = [h_1^{L}, h_2^L, \dots,h_N^{L}]$ as input and generates a sentence $\textbf{y} =[y_1,y_2,\dots, y_M]$ word-by-word, according to the hidden states of the topmost decoder layer $S^L = [s^L_1, s^L_2, \dots, s^L_M]$.

Formally, at time $t$, the $l$-th decoder layer ($l \in [1,\dots,L]$) updates the hidden state as:
\begin{equation}
\begin{split}
\hat{s}_t^{l} &= \text{SAN}(s_1^{l-1}, s_2^{l-1},\dots, s_{t}^{l-1}),\\
c_t^l &= \text{AN}(\hat{s}_t^{l}, H^{L}),	\\ 
s^{l}_t &= \text{FF}(c_t, \hat{s}_t^{l}), \\
\end{split}
\label{eq:dec}
\end{equation}
where $\text{FF}$ denotes a position-wise feed-forward layer, $[s_1^{l-1}, s_2^{l-1},\dots, s_{t}^{l-1}]$ represent the hidden states of the $l-1$th decoder layer, $[s^0_1, s^0_2,\dots, s^0_{t}]$ are embeddings of $[y_s,y_1,\dots,y_{t-2},y_{t-1}]$, and $y_s$ denotes the start symbol of a sentence.

In Eq~\ref{eq:dec}, $\text{AN}$ is a standard attention layer, which computes a set of attention scores $\beta_{ti} (i\in{[1,\dots,N]})$ and a context vector $c_t$:
\begin{equation}
\begin{split}
\beta_{ti} &= \frac{\exp(f({\hat{s}_t^l,h_i^L}))}{\sum\nolimits_{j \in [1,\dots, N]} \exp{(f({\hat{s}_t^l,h_j^L}))}}, \\
c^{l}_t &= \sum\nolimits_{i\in[1,\dots,N]} \beta_{ti} h_i^L,
\end{split}
\label{eq:dec-att}
\end{equation}
where $f$ is a scaled dot-product attention function. 

Denoting the output hidden state of the $L$-th decoder layer at time $t$ as $s_t^L$, the generator layer predicted the probability of a target word $y_t$ as:
\begin{equation}
p(y_t|\textbf{y}_{<t}, G) = \text{softmax}(W_gs_t^{L}),
\end{equation}
where $\textbf{y}_{<t} =  [y_1, y_2, \dots, y_{t-1}]$, and $W_g$ is a model parameter.

\subsection{Training Objective}
The training objective of the baseline model is to minimize the negative log-likelihood of conditional word probabilities:
\begin{equation}
\begin{split}
\ell_{std} &= -\sum\nolimits_{t\in[1,\dots,M]}log~p(y_t|\textbf{y}_{<t}, G) \\
& = -\sum\nolimits_{t\in[1,\dots,M]} log~p({y_t|s_t^L;\Theta}), \\
\end{split}
\label{eq:nmtloss}
\end{equation}
where $\Theta$ denotes the full set of parameters.

\section{Model with Back-Parsing}
Figure~\ref{fig:model} illustrates the proposed model. 
We adopt the baseline graph encoder described in Section~\ref{sec:graphencoder} for AMR encoding,
while enhancing the baseline decoder (Section~\ref{sec:stddecoder}) with  AMR graph prediction for better structure preservation.
In particular, we train the decoder to reconstruct the AMR graph (so called ``back-parsing'') by jointly predicting the corresponding AMR nodes and {\it projected} relations when generating a new word. 
In this way, we expect that the model can better memorize the AMR graph and generate more faithful outputs.
In addition, our decoder is trained in an \textit{online} manner, which uses the last node and edge predictions to better inform the generation of the next word.

Specifically, the encoder hidden states are first calculated given an AMR graph.
At each decoding time step, the proposed decoder takes the encoder states
as inputs and generates a new word (as in Section~\ref{sec:stddecoder}), together with its corresponding AMR node (Section~\ref{sec:nodeprediction}) and its outgoing edges (Section~\ref{sec:edgeprediction}), 
These predictions are then used inputs to calculate the next state (Section~\ref{sec:decinput}).

\subsection{Node Prediction}
\label{sec:nodeprediction}
We first equip a standard decoder with the ability to make word-to-node alignments while generating target words.
Making alignments can be formalized as a matching problem, which aims to find the most relevant AMR graph node for each target word.
Inspired by previous work~\cite{liu-etal-2016-neural,mi-etal-2016-supervised}, we solve the matching problem by supervising the word-to-node attention scores given by the Transformer decoder.
In order to deal with words without alignments, we introduce a {\it NULL} node $v_{\emptyset}$ into the input AMR graph (as shown in Figure~\ref{fig:model}) and align such words to it.\footnote{This node is set as the parent of the original graph root (e.g. {\it possible-01} in Figure~\ref{fig:model}) with relation ``\emph{root}''.}

More specifically, at each decoding step $t$, our Transformer decoder first calculates the top decoder layer word-to-node attention distribution $\beta_{t}'=[\beta_{t0}',\beta_{t1}',...,\beta_{tN}']$ (Eq~\ref{eq:dec} and Eq~\ref{eq:dec-att}) after taking the encoder states $H^L=[h_0^L, h_1^L, h_2^L, \dots, h_N^L]$ together with the previously generated sequence $\textbf{y}_{<t} =[y_1, y_2,\dots,y_{t-1}]$ ($\beta_{t0}'$ and $h_0^L$ are the probability and encoder state for the {\it NULL} node $v_{\emptyset}$).
Then the probability of aligning the current decoder state to node $v_i$ is defined as:
\begin{equation}
\begin{split}
p(\texttt{ALI}(s_t) = v_i| H^L, \textbf{y}_{<t}) = \beta_{ti}', \\
\end{split}
\label{eq:nodeprobability}
\end{equation}
where \texttt{ALI} is the sub-network for finding the best aligned AMR node for a given decoder state.

\noindent\textbf{Training.}
Supposing that the gold alignment (refer to Section~\ref{sec:expsetting}) at time $t$ is $\hat{\beta}_t$, the training objective for node prediction is to minimize the loss defined as the distance between $\beta_{t}'$ and $\hat{\beta}_t$:
\begin{equation}
\begin{split}
\ell_{node} = \sum\nolimits_{t\in[1,\dots,M]} \Delta(\beta_{t}', \hat{\beta_t}), \\
\end{split}
\label{eq:nodeloss}
\end{equation}
where $\Delta$ denotes a discrepancy criterion that can quantify the distance between $\beta_{t}'$ and $\hat{\beta_t}$.
We take two common alternatives: (1) Mean Squared Error (MSE), 
and (2) Cross Entropy Loss (CE). 

\subsection{Edge Prediction}
\label{sec:edgeprediction}
The edge prediction sub-task aims to preserve the node-to-node relations in an AMR graph during text generation.
To this end, we project the edges of each input AMR graph onto the corresponding sentence according to their node-to-word alignments, before training the decoder to generate the \textit{projected} edges along with target words.
For words without outgoing edges, we add a ``self-loop'' edge for consistency.

Formally, at decoding step $t$, each relevant directed edge (or \textit{arc}) with relation label $r_k$ starting from $y_t$ can be represented as $\left\langle y_j, r_k, y_t\right\rangle$, where $j\leq t$, $y_j$, $y_t$ and  $r_k$ are called ``{\it arc\_to}'', ``{\it arc\_from}'', and ``{\it label}'' respectively.
We modify the deep biaffine attention classifier \cite{dozat2016deep} to model these edges. 
In particular, we factorize the probability for each labeled edge into the ``{\it arc}'' and ``{\it label}'' parts, computing both based on the current decoder hidden state and the states of all previous words.
The ``{\it arc}'' score $\psi^\text{arc}_{tj} \in \mathcal{R}^{1}$, which measures whether or not a directed edge from $y_t$ to $y_j$ exists, is calculated as:
\begin{equation}
\begin{split}
&b_j^{\text{arc\_to}}, b_t^\text{arc\_from} = \text{FF}^{\text{arc\_to}}(s_j^L), \text{FF}^{\text{arc\_from}}(s_t^L), \\
&\hat{\psi}^\text{arc}_{tj} = \text{Biaff}^\text{arc}(b_j^\text{arc\_to}, b_t^\text{arc\_from}), \\
&\psi^\text{arc}_{t1}, \psi^\text{arc}_{t2}, ..., \psi^\text{arc}_{tj},...,\psi^\text{arc}_{tt} \\
&\quad~~= \text{softmax}(\hat{\psi}^\text{arc}_{t1}, \hat{\psi}^\text{arc}_{t2}, ..., \hat{\psi}^\text{arc}_{tj},...,\hat{\psi}^\text{arc}_{tt}). \\
\end{split}
\label{eq:biaffine-arc}
\end{equation}
Similarly, the ``{\it label}'' score $\psi^\text{label}_{tj}\in \mathcal{R}^{R}$, which is used to predict a label for potential word pair ($y_j$, $y_t$), is given by:
\begin{equation}
\begin{split}
b_j^\text{label\_to}, &b_t^\text{label\_from} = \text{FF}^{\text{label\_to}}(s_j^L), \text{FF}^{\text{label\_from}}(s_t^L), \\
\psi^\text{label}_{tj} &= \text{softmax}\big(\text{Biaff}^\text{label}(b_j^\text{label\_to}, b_t^\text{label\_from})\big). \\
\end{split}
\label{eq:biaffine-label}
\end{equation}

In Eq~\ref{eq:biaffine-arc} and Eq~\ref{eq:biaffine-label}, $ \text{FF}^{\text{arc\_to}}$, $ \text{FF}^{\text{arc\_from}}$, $\text{FF}^{\text{label\_to}}$ and $\text{FF}^{\text{label\_from}}$ are linear transformations. $\text{Biaff}^\text{arc}$ and $\text{Biaff}^\text{label}$ are biaffine transformations:
\begin{equation}
\begin{split}
\text{Biaff}(x_1, x_2) = x_1^T U x_2 + W (x_1 \oplus x_2) + b,
\end{split}
\label{eq:biaffine}
\end{equation}
where $\oplus$ denotes vector concatenation, $U, W$ and $b$ are model parameters. 
$U$ is a $(d \times 1 \times d)$ tensor for unlabeled classification (Eq~\ref{eq:biaffine-arc}) and a $(d \times R \times d)$ tensor for labeled classification (Eq~\ref{eq:biaffine-label}), where $d$ is the hidden size.

Defining $p(y_j|y_t)$ as $\psi^\text{arc}_{tj}$ and $p(r_k|y_j,y_t)$ as $\psi^\text{label}_{tj}[k]$,
the probability of a labeled edge $\left\langle y_j, r_k, y_t\right\rangle$ is calculated by the chain rule:
\begin{equation}
\begin{split}
p(r_k,y_j|y_t) & = p(r_k|y_j,y_t) p(y_j|y_t) \\ 
& = \psi^\text{label}_{tj}[k] \cdot \psi^\text{arc}_{tj}. \\
\end{split}
\label{eq:edgep}
\end{equation}

\noindent\textbf{Training.}
The training objective for the edge prediction task is the negative log-likelihood over all {\it projected} edges ${E'}$:
\begin{equation}
\begin{split}
\ell_{label} &= -\sum\nolimits_{	\left \langle y_j, r_{k},y_i  \right \rangle \in E'}log~p(r_k,y_j|y_{i}) \\
\end{split}
\end{equation}

\subsection{Next State Calculation}
\label{sec:decinput}
In addition to simple ``one-way'' AMR back-parsing (as shown in Section \ref{sec:nodeprediction} and \ref{sec:edgeprediction}), we also study integrating the previously predicted AMR nodes and outgoing edges as additional decoder inputs to help generate the next word. 
In particular, for calculating the decoder hidden states $[s^1_{t+1}, s^2_{t+1},...,s^L_{t+1}]$ at step $t+1$, the input feature to our decoder is a triple $\left\langle \vec{y}_{t}, \vec{v}_{t}, \vec{e}_{t} \right\rangle$ instead of a single value $\vec{y}_t$, which the baseline has. Here $\vec{y}_{t}, \vec{v}_{t}$ and $\vec{e}_{t}$ are vector representations of the predicted word, AMR node and edges at step $t$, respectively. 
More specifically, $\vec{v}_t$ is a weighted sum of the top-layer encoder hidden states $[h_0^L, h_1^L, ...,h_{N}^L]$, and coefficients are from the distribution of $\beta^\prime_t$ in Eq~\ref{eq:nodeprobability}:
\begin{equation}
\begin{split}
\vec{v}_{t} &= \sum\nolimits_{i\in[0,\dots,N]} \beta^\prime_{ti} \odot h_i^L, \\
\end{split}
\label{eq:nodeavg}
\end{equation}
where $\odot$ is the operation for scalar-tensor product.

Similarly, $\vec{e}_{t}$ is calculated as:
\begin{equation}
\begin{split}
\vec{e}_{t} &= \vec{r}_{t}\oplus \vec{s}_{t}, \\
\vec{r}_{t} &= \sum\nolimits_{k=1}^{|R|}\sum\nolimits_{j=1}^{t} p(r_k,y_j|y_t) \gamma_{k}, \\
\vec{s}_{t} &= \sum\nolimits_{j=1}^{t} p(y_j|y_t) s^L_j,\\
\end{split}
\label{eq:edgeavg}
\end{equation}
where $\oplus$ concatenates two tensors, $p(r_k,y_j|y_t)$ and $p(y_j|y_t)$ are probabilities given in Eq~\ref{eq:edgep},
$\gamma_{k}$ is a relation embedding, and $s^L_j$ is the decoder hidden state at step $j$.
$\vec{e}_{t-1}$ is a vector concatenation of $\vec{r}_{t}$ and $\vec{s}_{t}$, which are weighted sum of relation embeddings and weighted sum of previous decoder hidden states, respectively.

In contrast to the baseline in Eq~\ref{eq:dec}, at time $\textit{\textbf{t+1}}$, the hidden state of the \textbf{first} decoder layer is calculated as: 
\begin{equation}
\begin{split}
\hat{s}_{t+1}^{1} &= \text{SAN}(s_1^{0},...,s_{t}^{0}, \vec{y}_{t}, \vec{v}_{t}, \vec{e}_{t}),\\
c_{t+1}^1 &= \text{AN}(\hat{s}_{t+1}^{1}, H^{L}),	\\ 
s^{1}_{t+1} &= \text{FF}(c_{t+1}^1, \hat{s}_{t+1}^{1}), \\
\end{split}
\label{eq:decours}
\end{equation}
where the definition of $H^L$, SAN, AN, FF and $[s^0_1,\dots,s^0_{t}]$ are the same as Eq~\ref{eq:dec}.
$\vec{v}_{0}$ and $\vec{e}_{0}$ (as shown in Figure \ref{fig:model}) are defined as zero vectors.
The hidden states of upper decoder layers ($[s^2_{t+1},...,s^L_{t+1}]$) are updated in the same way as Eq~\ref{eq:dec}.


Following previous work on syntactic text generation \cite{wu2017sequence,wang2018tree}, we use gold AMR nodes and outgoing edges as inputs for training, while we take automatic predictions for decoding.

\subsection{Training Objective}
The overall training objective is:
\begin{equation}
\begin{split}
\ell_{total} &=  \ell_{std} + \lambda_1\ell_{node} + \lambda_2\ell_{label},\\
\end{split}
\end{equation}
where $ \lambda_1$ and $ \lambda_2$ are weighting hyper-parameters
for $\ell_{node}$ and $\ell_{label}$, respectively.

\section{Experiments}
We conduct experiments on two benchmark AMR-to-text generation datasets, including LDC2015E86 and LDC2017T10.
These two datasets contain 16,833 and 36,521 training examples, respectively, and share a common set of 1,368 development and 1,371 test instances.

\subsection{Experimental Settings}
\label{sec:expsetting}

\noindent \textbf{Data preprocessing.} Following previous work \cite{song2018graph,zhu2019modeling}, we take a standard simplifier \cite{konstas2017neural} to preprocess AMR graphs, adopting the Stanford tokenizer\footnote{https://nlp.stanford.edu/software/tokenizer.shtml} and Subword Tool\footnote{https://github.com/rsennrich/subword-nmt} to segment text into subword units. 
The node-to-word alignments are generated by ISI aligner \cite{pourdamghani-etal-2014-aligning}. 
We then project the source AMR graph onto the target sentence according to such alignments.

For node prediction, the attention distributions are normalized, but the alignment scores generated by the ISI aligner are unnormalized hard $0/1$ values.
To enable cross entropy loss, we follow previous work \cite{mi-etal-2016-supervised} to normalize the gold-standard alignment scores.

\noindent \textbf{Hyperparameters.} 
We choose the feature-based model\footnote{We do not choose their best model (G-Trans-SA) due to its large GPU memory consumption, and its performance is actually comparable with G-Trans-F in our experiments.} of  \citet{zhu2019modeling} as our baseline (G-Trans-F-Ours).
Also following their settings, both the encoder and decoder have $6$ layers, with each layer having $8$ attention heads.
The sizes of hidden layers and word embeddings are $512$, and the size of relation embedding is $64$.
The hidden size of the biaffine attention module is 512.
We use Adam~\cite{Kingma2015AdamAM} with a learning rate of 0.5 for optimization. 
Our models are trained for 500K steps on a single 2080Ti GPU.
We tune these hyperparameters on the LDC2015E86 development set and use the selected values for testing\footnote{Table~\ref{tab:allparam} in Appendix shows the full set of parameters.}.

\noindent \textbf{Model Evaluation.} We set the decoding beam size as $5$ and take BLEU~\cite{papineni-2002-bleu} and Meteor~\cite{banerjee-lavie-2005-meteor,denkowski-lavie-2014-meteor} as automatic evaluation metrics. 
We also employ human evaluation to assess the semantic faithfulness and generation fluency of compared methods by randomly selecting 50 AMR graphs for comparison. 
Three people familiar with AMR are asked to score the generation quality with regard to three aspects --- concept preservation rate, relation preservation rate and fluency (on a scale of [0, 5]). Details about the criteria are: 

\begin{adjustwidth}{0.5cm}{0cm}
	\indent\textbullet~Concept preservation rate assesses to what extent the concepts in input AMR graphs are involved in generated sentences. \\
	\noindent\textbullet~Relation preservation rate measures to what extent the relations in input AMR graphs exist in produced utterances. \\
	\noindent\textbullet~Fluency evaluates whether the generated sentence is fluent and grammatically correct. \\
\end{adjustwidth}

Recently, significant progress ~\cite{ribeiro2019enhancing,Zhang2020BERTScore,elikyilmaz2020EvaluationOT} in developing new metrics for NLG evaluation has made.
We leave evaluation on these metrics for future work.
\subsection{Development Experiments}
\label{sec:dev}
\begin{table}[!t]
	\small
	\centering{
		\begin{tabular}{lcc}
			\toprule
			\textbf{Model} & BLEU & Meteor \ \\
			\midrule
			G-Trans-F-Ours &30.20  &35.23 \\
			\midrule
			Node Prediction MSE  &30.66 &35.60 \\
			Node Prediction CE  &\textbf{30.85}  &\textbf{35.71} \\
			\midrule
			Edge Prediction share &\textbf{31.19}  &\textbf{35.75} \\
			Edge Prediction independent &31.13  &35.69 \\
			\bottomrule
		\end{tabular}
	}
	\caption{\label{tab:dev} BLEU and Meteor scores on the LDC2015E86 devset under different model settings.}
\end{table}

Table \ref{tab:dev} shows the performances on the devset of LDC2015E86 under different model settings. 
For the node prediction task, it can be observed that both  cross entropy loss (CE) and mean squared error loss (MSE) give significantly better results than the baseline, with 0.46 and 0.65 improvement in terms of BLEU, respectively.
In addition, CE gives a better result than MSE.

Regarding edge prediction, we investigate two settings, with relation embeddings being shared by the encoder and decoder, or being separately constructed, respectively.
Both settings give large improvements over the baseline.
Compared with the model using independent relation embeddings, the model with shared relation embeddings gives slightly better results with less parameters, indicating that the relations in an AMR graph and the relations between words are consistent.
We therefore adopt the CE loss and shared relation embeddings for the remaining experiments.
\begin{figure}
	\centering
	\includegraphics[width=0.99\linewidth]{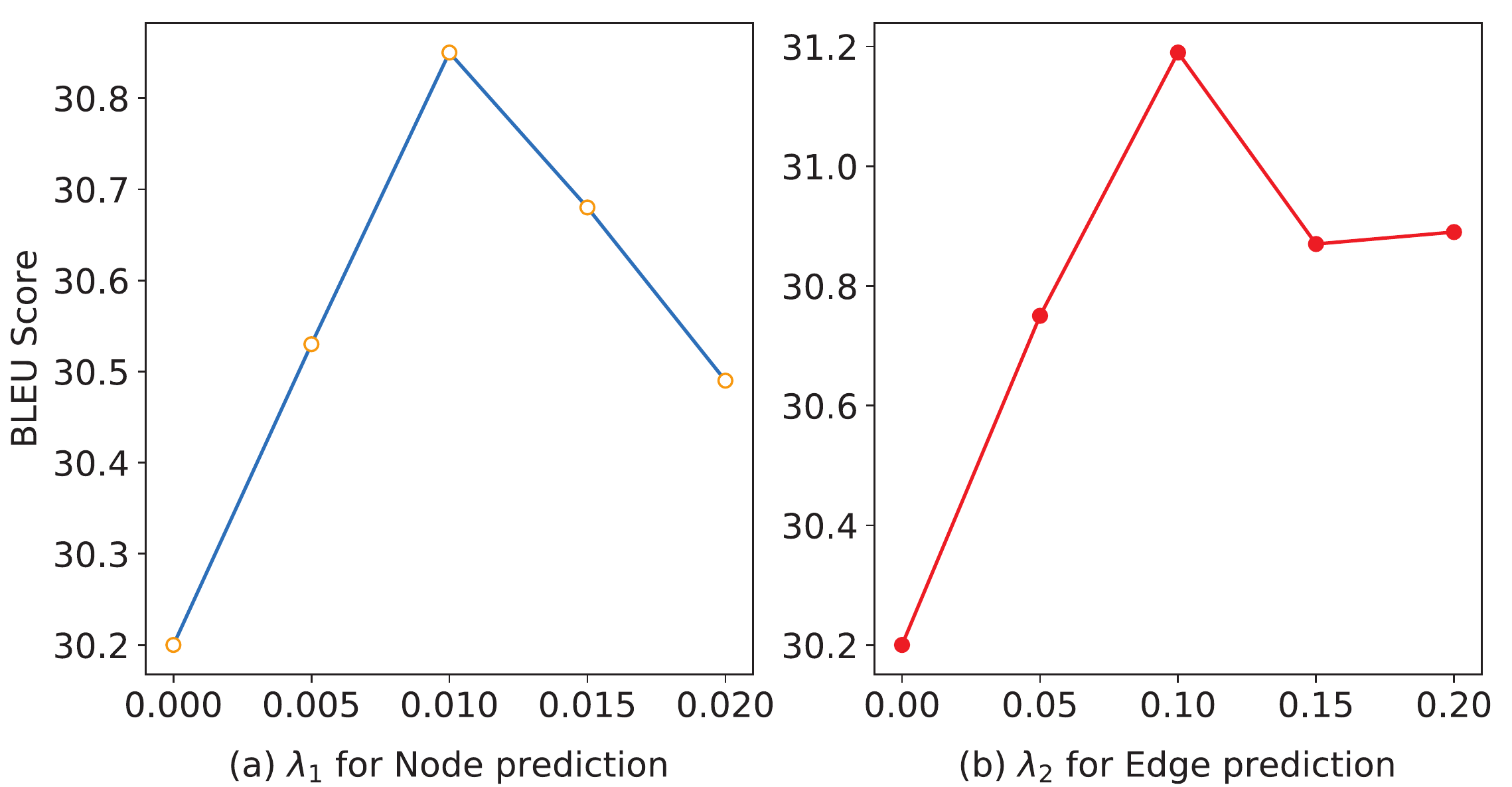}
	\caption{BLEU scores on the LDC2015E86 devset against different hyperparameter values.}
	\label{fig:dev}
\end{figure}

Figure~\ref{fig:dev} presents the BLEU scores of integrating standard AMR-to-text generation with node prediction or edge prediction under different $\lambda_1$ and $\lambda_2$ values, respectively. 
There are improvements when increasing the coefficient from $0$, demonstrating that both node prediction and edge prediction have positive influence on AMR-to-text generation. 
The BLEU of the two models reach peaks at $\lambda_1=0.01$ and $\lambda_2=0.1$, respectively. When further increasing the coefficients, the BLEU scores start to decrease.
We thus set $\lambda_1=0.01, \lambda_2=0.1$ for the rest of our experiments. 
\subsection{Main Results}
\subsubsection{Automatic Evaluation}
\begin{table}
	\small
	\begin{center}
		\resizebox{0.48\textwidth}{!}{
			\begin{tabular}{lcc}
				\toprule
				\multirow{1}{*}{\textbf{Model}}& \multicolumn{1}{c}{\textbf{LDC15}} &\multicolumn{1}{c}{\textbf{LDC17}} \\
				\midrule
				LSTM~\cite{konstas2017neural} &22.00 & -- \\
				GGNN~\cite{beck2018graph} & -- & 23.30 \\
				GRN~\cite{song2018graph} &23.30 & -- \\
				DCGCN~\cite{guo-etal-2019-densely} &25.9 &27.9  \\
				G-Trans-F~\cite{zhu2019modeling} &27.23  &30.18   \\
				G-Trans-SA~\cite{zhu2019modeling} &29.66  &31.54  \\
				G-Trans-C~\cite{cai2020graph} &27.4  &29.8   \\
				G-Trans-W~\cite{wang2020amr} &25.9  &29.3   \\
				\midrule
				G-Trans-F-Ours &30.15  &31.53   \\
				Ours Back-Parsing &\textbf{31.48} &\textbf{34.19} \\
				\midrule
				\textit{with external data} & & \\
				LSTM (20M)~\cite{konstas2017neural} &33.8  & -   \\
				GRN (2M)~\cite{song2018graph} &33.6  &-  \\
				G-Trans-W (2M)~\cite{wang2020amr} &36.4  &- \\
				\bottomrule
			\end{tabular}}
			\caption{Test-set BLEU scores on LDC2015E86 (LDC15) and LDC2017T10 (LDC17).}
			\label{tab:main}
	\end{center}
\end{table}
Table~\ref{tab:main} shows the automatic evaluation results, where ``\emph{G-Trans-F-Ours}'' and ``\emph{Ours Back-Parsing}'' represent the baseline and our full model, respectively.
The top group of the table shows the previous state-of-the-art results on the LDC2015E86 and LDC2017T10 testsets.
Our systems give significantly better results than the previous systems using different encoders, including 
LSTM~\cite{konstas2017neural}, 
graph gated neural network (GGNN; \citealp{beck2018graph}), graph recurrent network (GRN;~\citealp{song2018graph}), densely connected graph convolutional network (DCGCN;~\citealp{guo-etal-2019-densely}) and various graph transformers (G-Trans-F, G-Trans-SA, G-Trans-C, G-Trans-W). 
Our baseline also achieves better BLEU scores than the corresponding models of \citet{zhu2019modeling}.
The main reason is that we train with more steps (500K vs 300K) and we do not prune low-frequency vocabulary items after applying BPE. 
Note that we do not compare our model with methods by using external data.

Compared with our baseline (G-Trans-F-Ours), 
the proposed approach achieves significant ($p<0.01$) improvements, giving BLEU scores of $31.48$ and $34.19$ on LDC2015E86 and LDC2017T10, respectively, which are to our knowledge the best reported results in the literature.
In addition, the outputs of our model have $0.8$ more words than the baseline on average.
Since the BLEU metric tend to prefer shorter results, this confirm that our model indeed recovers more information.

\subsubsection{Human Evaluation}
As shown in Table~\ref{tab:human}, our model gives higher scores of concept preservation rate than the baseline on both datasets, with improvements of 3.6 and 3.3, respectively. 
In addition, the relation preservation rate of our model is also better than the baseline.
This indicating that our model can preserve more concepts and relations than the baseline method, thanks to the back-parsing mechanism. 
With regard to the generation fluency, our model also gives better results than baseline.
The main reason is that the relations between concepts such as \textit{subject-predicate relation} and \textit{modified relation} are helpful for generating fluency sentences.
\begin{table}
	\small
	\centering{
			\begin{tabular}{lccc}
				\toprule
				\textbf{Setting}& \textbf{CPR(\%)} & \textbf{RPR(\%)} & \textbf{Fluency} \\
				\midrule
				\textit{LDC2015E86} & & & \\
				Baseline & {92.19} &{88.79} &{4.08} \\
				Ours & \textbf{95.80} & \textbf{91.33} &\textbf{4.34} \\
				\midrule
				\textit{LDC2017T10} & & & \\
				Baseline & {93.36} &{90.05} &{4.15} \\
				Ours & \textbf{96.63} &\textbf{92.21} &\textbf{4.42} \\
				\bottomrule
			\end{tabular}
	}
	\caption{\label{tab:human} Human evaluation of the sentences generated by different systems on concept presevation rate (CPR), relation preservation rate (RPR) and fluency.}
\end{table}
\begin{table}
	\small
	\centering{
		\begin{tabular}{lcccc}
			\toprule
			\textbf{Model}& Cause & Contrast & Condition  & Coord. \\
			\midrule
			Baseline &0.84  &0.92 &0.91 &\textbf{0.98} \\
			Ours & \textbf{0.96} &\textbf{0.98} &\textbf{0.95} &\textbf{0.98} \\
			\bottomrule
		\end{tabular}
	}
	\caption{\label{tab:discourse} Human study for discourse preservation accuracy on LDC2015E86.} 
\end{table}
Apart from that, we study discourse~\cite{prasad-etal-2008-penn}  relations, which are essential for generating a good sentence with correct meaning.
Specifically, we consider $4$ common discourse relations (``Cause'', ``Contrast'', ``Condition'', ``Coordinating'').  For each type of discourse, we randomly select 50 examples from the test set and ask 3 linguistic experts to calculate the discourse preservation accuracy by checking if the generated sentence preserves such information.

Table~\ref{tab:discourse} gives discourse preservation accuracy results of the baseline and our model, respectively.
The baseline already performs well, which is likely because discourse information can somehow be captured through co-occurrence in each (AMR, sentence) pair.
Nevertheless, our approach achieves better results, showing that our back-parsing mechanism is helpful for preserving discourse relations.
\subsection{Analysis}

\paragraph{Ablation}
We conduct ablation tests to study the contribution of each component to the proposed model.
In particular, we evaluate models with only the node prediction loss (Node Prediction, Section~\ref{sec:nodeprediction}) and the edge prediction loss (Edge Prediction, Section~\ref{sec:edgeprediction}), respectively, and
further investigate the effect of integrating node and edge information into the next state computation (Section~\ref{sec:decinput}) by comparing models without and with (Int.) such integration.
\begin{table}
	\small
	\centering{
		\begin{tabular}{lcc}
			\toprule
			\textbf{Model}& BLEU & Meteor\\
			\midrule
			Baseline &30.15  &35.36 \\
			\midrule
			\quad + Node Prediction  &30.49  &35.66 \\
			\quad + Node Prediction (Int.) & 30.72 &35.94  \\
			\midrule
			\quad + Edge Prediction  &30.80  &35.71 \\
			\quad + Edge Prediction (Int.) & 31.07 &35.87 \\
			\midrule
			\quad + Both Prediction  &30.96 &35.92 \\
			\quad + Both Prediction (Int.) & \textbf{31.48} &\textbf{36.15} \\
			\bottomrule
		\end{tabular}
	}
	\caption{\label{tab:ablation} Ablation study on LDC2015E86 test set.}
\end{table}

Table~\ref{tab:ablation} shows the BLEU and Meteor scores on the LDC2015E86 testset.
Compared with the baseline, we observe a performance improvement of 0.34 BLEU  by adding the node prediction loss only.
When using the predicted AMR graph nodes as additional input for next state computation (i.e., Node Prediction (Int.)), the BLEU score increases from 30.49 to 30.72, and the Meteor score reaches 35.94, showing that the previously predicted nodes are beneficial for text generation.
Such results are consistent with our expectation that predicting the corresponding AMR node can help the generation of correct content words (a.k.a. concepts).

Similarly, edge prediction also leads to performance boosts.
In particular, integrating the predicted relations for next state computation (Edge Prediction (Int.)) gives an improvement of 0.92 BLEU over the baseline.
Edge prediction results in larger improvements than node prediction, indicating that relation knowledge is more informative than word-to-node alignment.

In addition, combining the node prediction and edge prediction losses (Both Prediction) leads to better model performance, which indicates that node prediction and edge prediction have mutual benefit.
Integrating both node and edge predictions (Both Prediction (Int.)) further improves the system to 31.48 BLEU and 36.15 Meteor, respectively.

\begin{figure}
	\centering
	\includegraphics[width=0.99\linewidth]{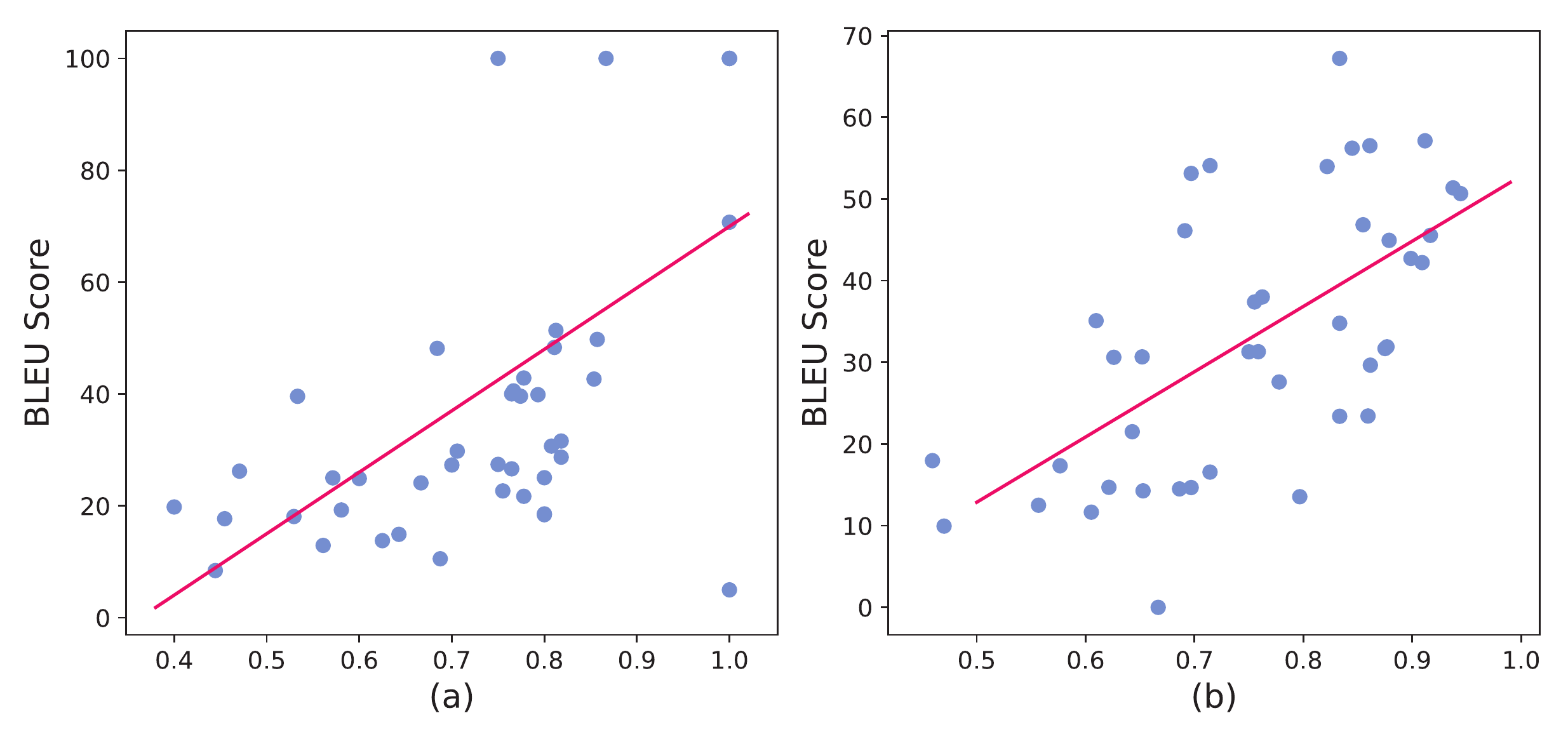}
	\caption{Performance (in BLEU) on the test set with respect to the node (a) and edge (b) prediction accuracy.}
	\label{fig:correlation}
\end{figure}

\begin{table}
	\small
	\centering{
		\begin{tabular}{lcc}
			\toprule
			\textbf{Setting}& \textbf{LDC2015E86} & \textbf{LDC2017T10}  \\
			\midrule
			Node Prediction Acc. & {0.65} &{0.71}  \\
			Edge Prediction Acc. & {0.56} &{0.59}  \\
			Both Prediction Acc. & \textbf{0.69} & \textbf{0.73}  \\
			\bottomrule
		\end{tabular}
	}
	\caption{\label{tab:pearson} The pearson correlation coefficient $\rho$ between the prediction accuracy and BLEU.}
\end{table}

\noindent \textbf{Correlation between Prediction Accuracy and Model Performance}~~We further investigate the influence of AMR-structure preservation on the performance of the main text generation task.
Specifically, we first force our model to generate a gold sentence in order to calculate the accuracies for node prediction and edge prediction.
We then calculate the corresponding BLEU score for the sentence generated by our model on the same input AMR graph without forced decoding, before drawing correlation between the accuracies and the BLEU score.
As shown in Figure \ref{fig:correlation}(a) and \ref{fig:correlation}(b)\footnote{For clear visualization, we only select the first one out of every 30 sentences from the LDC2015E86 testset.}, both node accuracy and edge accuracy have a strong positive correlation with the BLEU score, indicating that the more structural information is retained, the better the generated text is.

We also evaluate the pearson ($\rho$) correlation coefficients between BLEU scores and node (edge) prediction accuracies.
Results are given in Table~\ref{tab:pearson}.
Both types of prediction accuracies have strong positive correlations with the final BLEU scores, and their 
combination yields further boost on the correlation coefficient, indicating the necessity of jointly predicting the nodes and edges.

\begin{figure}
	\centering
	\includegraphics[width=0.8\linewidth]{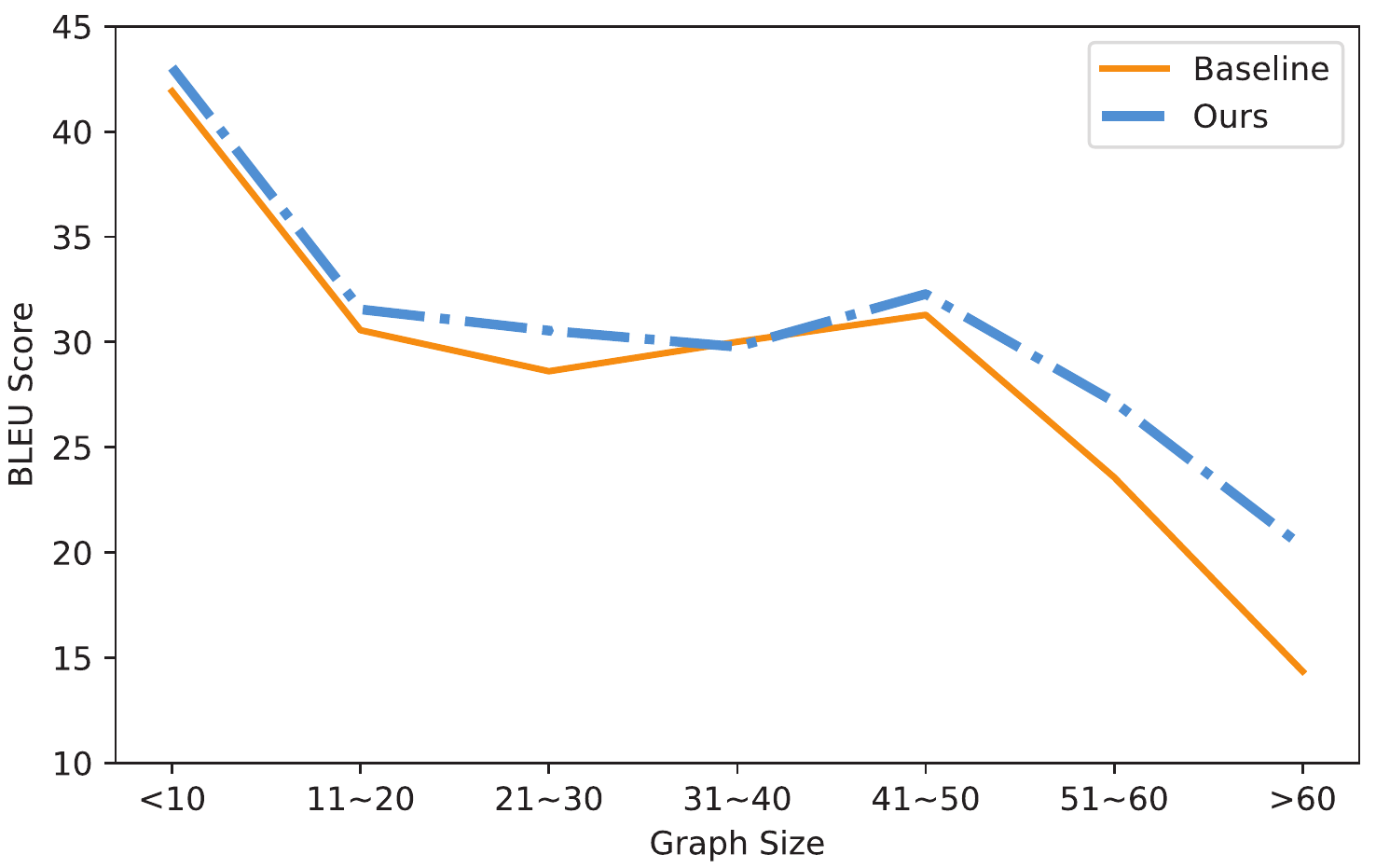}
	\caption{Performances (in BLEU) on the test set with respect to the size of the input AMR graphs.}
	\label{fig:graphsize}
\end{figure}

\noindent\textbf{Performances VS AMR Graphs Sizes}
Figure \ref{fig:graphsize} compares the BLEU scores of the baseline and our model on different AMR sizes.
Our model is consistently better than the baseline for most length brackets, and the advantage is more obvious for large AMRs (size 51+).

\begin{figure}
	\centering
	\includegraphics[width=0.99\linewidth]{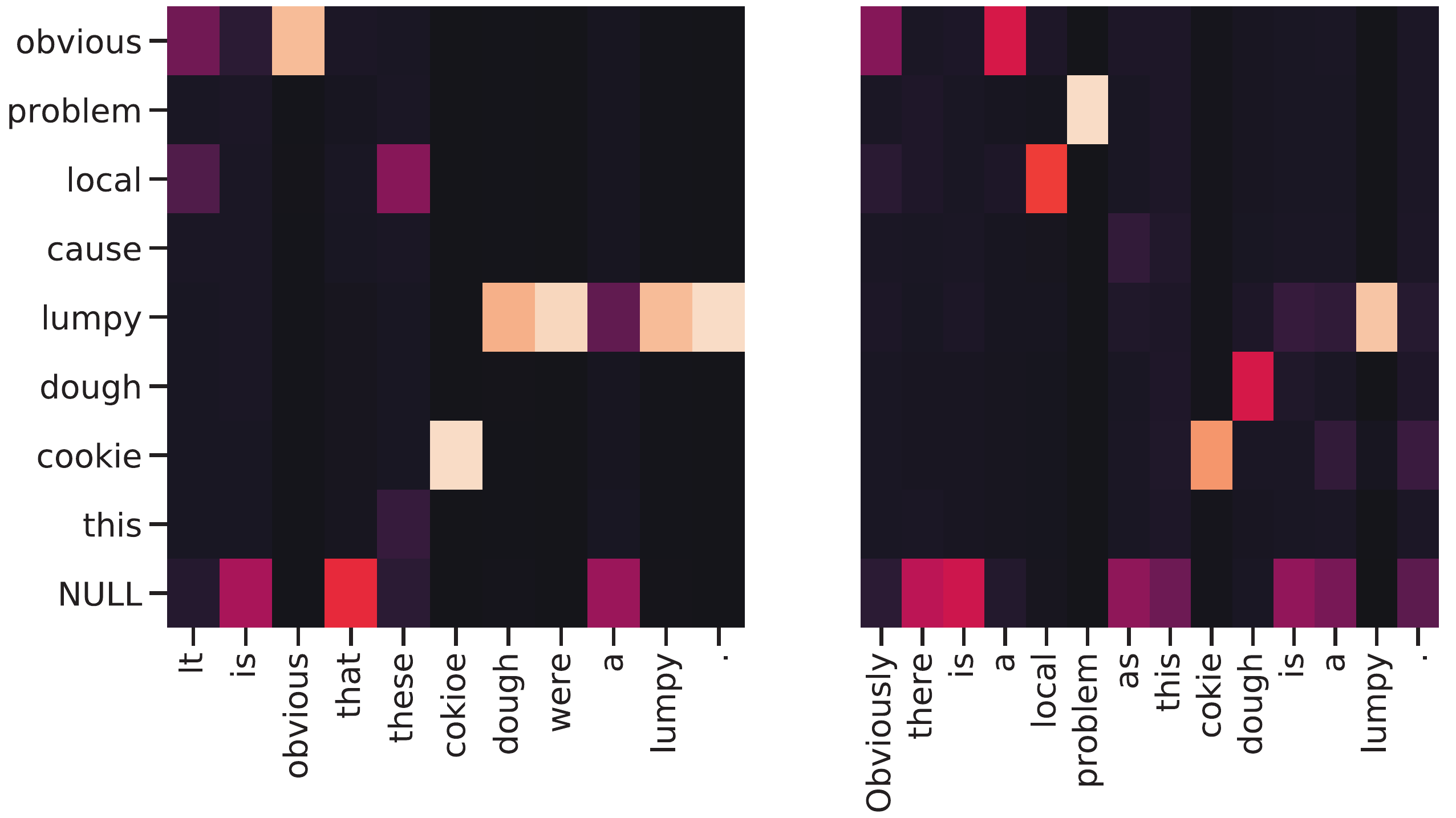}
	\caption{Visualization of word-to-node attention obtained from the baseline graph Transformer (left) and our model with node prediction loss (right).}
	\label{fig:attn}
\end{figure}

\subsection{Case Study}
We provide two examples in Table~\ref{tab:case} to help better understand the proposed model. 
Each example consists of an AMR graph, a reference sentence (\textbf{REF}), the output of baseline model (\textbf{Baseline}) and the sentence generated by our method (\textbf{Ours}).

As shown in the first example, although the baseline model maintains the main idea of the original text, it fails to recognize the AMR graph nodes {\it ``local''} and {\it ``problem''}. 
In contrast, our model successfully recovers these two nodes and generates a sentence which is more faithful to the reference. 
We attribute this improvement to node prediction. 
To verify this, we visualize the word-to-node attention scores of both approaches in Figure~\ref{fig:attn}. 
As shown in the figure, the baseline model gives little attention to the AMR node \textit{``local''} and \textit{``problem''} during text generation.
In contrast, our system gives a more accurate alignment to the relevant AMR nodes in decoding.

In the second example, the baseline model incorrectly positions the terms \textit{``doctor''}, \textit{``see''} and \textit{``worse cases''} while our approach generates a more natural sentence. This can be attributed to the edge prediction task, which can inform the decoder to preserve the relation that \textit{``doctor''} is the subject of \textit{``see''} and \textit{``worse cases''} is the object.

\begin{table}[!htbp]
	\small
	\centering{
		\begin{tabular}{l}
			\toprule
			\textbf{(1)}  (o / obvious-01 \\ 
			\qquad :ARG1 (p / problem \\
			\qquad \qquad:ARG1-of (l / local-02)) \\ 
			\qquad :ARG1-of (c / cause-01 \\
			\qquad \qquad:ARG0 (l2 / lumpy \\
			\qquad \qquad \qquad:domain (d / dough \\
			\qquad \qquad \qquad \qquad:mod (c2 / cookie) \\
			\qquad \qquad \qquad \qquad:mod (t / this)))))\\
			\textbf{REF}: Obviously there are local problems because this\\
			cookie dough is lumpy .\\
			\textbf{Baseline}:  It is obvious that these cookie dough were a\\   lumpy . \\
			\textbf{Ours}: Obviously there is a {\color{blue} local problem} {\color{green} as } this cookie \\ dough is a lumpy . \\
			\midrule
			\textbf{(2)} (c / cause-01 \\
			\qquad :ARG0 (s / see-01 \\
			\qquad\qquad :ARG0 (d / doctor) \\
			\qquad\qquad :ARG1 (c2 / case \\
			\qquad\qquad\qquad:ARG1-of (b / bad-05 \\
			\qquad\qquad\qquad\qquad:degree (m / more \\
			\qquad\qquad\qquad\qquad\qquad:quant (m2 / much))))) \\
			\qquad:ARG1 (w / worry :polarity - :mode imperative \\
			\qquad\qquad:ARG0 (y / you) \\
			\qquad\qquad:ARG1 (t / that))) \\
			\textbf{REF}: Doctors have seen much worse cases so don't \\ worry about that ! \\ 
			\textbf{Baseline}: Don't worry about that {\color{red} see much worse} \\
			{\color{red} cases by doctors} . \\
			\textbf{Ours}: Don't worry that , {\color{green} as } a {\color{blue} doctor saw much worse} \\ 
			{\color{blue} cases} . \\
			\bottomrule
		\end{tabular}
	}
	\caption{\label{tab:case} Examples for case study.}
\end{table}


\section{Related Work}

Early studies on AMR-to-text generation rely on statistical methods. \citet{flanigan2016generation} convert input AMR graphs to trees by splitting re-entrances, before translating these trees into target sentences with a tree-to-string  transducer;
\citet{pourdamghani2016generating} apply a phrase-based MT system on linearized AMRs; \citet{song2017amr} design a synchronous node replacement grammar to parse input AMRs while generating target sentences.
These approaches show comparable or better results than early neural models \cite{konstas2017neural}.
However, recent neural approaches \cite{song2018graph, zhu2019modeling,cai2020graph,wang2020amr,mager-etal-2020-gpt} have demonstrated the state-of-the-art  performances thanks to the use of contextualized embeddings.


Related work on NMT studies back-translation loss~\cite{sennrich2016improving,tu2017neural} by translating the target reference back into the source text (reconstruction), which can help retain more comprehensive input information. This is similar to our goal.
\citet{wiseman2017challenges} extended 
the reconstruction loss of 
\citet{tu2017neural} for table-to-text generation.
We study a more challenging topic on how to retain the meaning of a complex graph structure rather than a sentence or a table.
In addition, rather than reconstructing the input \textit{after} the output is produced, we predict the input \textit{while} the output is constructed, thereby allowing stronger information sharing.


Our work is also remotely related to previous work on string-to-tree neural machine translation (NMT) \citep{aharoni2017towards,wu2017sequence,wang2018tree}, which aims at generating target sentences together with their syntactic trees.
One major difference is that their goal is producing grammatical outputs, while ours is preserving input structural information.

\section{Conclusion}

We investigated back-parsing for AMR-to-text generation by integrating the prediction of projected AMRs into sentence decoding.
The resulting model benefits from both richer loss and more structual features during decoding.
Experiments on two benchmarks show advantage of our model over a state-of-the-art baseline.
\section*{Acknowledgments}
Yue Zhang is the corresponding author.
We would like to thank the anonymous reviewers for their insightful comments and Yulong Chen for his fruitful inspiration.
This work has been supported by National Natural Science Foundation of China under grant No.61976180 and a Xiniuniao grant of Tencent.

\bibliographystyle{acl_natbib}
\bibliography{emnlp2020}

\clearpage
\appendix
\section{Appendices}
\subsection{Full Experimental Settings}
\begin{table}
	\small
	\centering{
		\begin{tabular}{ll|ll}
			\toprule
			Parameter& Value & Parameter & Value \\
			\midrule
			Src vocab (BPE) &10,004  & Optimizer &Adam \\
			Tgt vocab (BPE) &10,004  & Learning rate &0.5 \\
			Relation vocab &5,002  &Adam beta$_1$  &0.9 \\		
			Encoder layer &6 &Adam beta$_2$  &0.98 \\
			Decoder layer &6 &Lr decay & 0.5 \\
			Hidden size &512 &Decay method & noam \\
			Attention heads &8 &Decay step & 10,000 \\
			Attention dropout &0.3 &Warmup &16,000 \\
			Share\_embeddings &True &Batch size &2048 \\
			Python version &3.6 & $\lambda_1$ &0.01\\
			Pytorch version &1.0.1 & $\lambda_2$ &0.1\\
			Model parameters &67.93M & Training time &30h\\
			\bottomrule
		\end{tabular}
	}
	\caption{\label{tab:allparam} Full list of model parameters on the LDC2015E86.} 
\end{table}
\begin{table*}[!hbp]
	\small
	\centering{
		\begin{tabular}{lcccccc}
			\toprule
			\multirow{2}{*}{\textbf{Model}}& \multicolumn{3}{c}{\textbf{LDC2015E86}} &\multicolumn{3}{c}{\textbf{LDC2017T10}} \\
			\cmidrule(lr){2-4} \cmidrule(lr){5-7}
			&BLEU &Meteor &CHRF++  &BLEU & Meteor &CHRF++ \\
			\midrule
			LSTM~\cite{konstas2017neural} &22.00 &- &- &-  &- &- \\
			GRN~\cite{song2018graph} &23.30 &- &- &-  &- &- \\
			Syntax-G~\cite{cao-clark-2019-factorising} &23.5 &- & - &26.8 & - & - \\
			S-Enc~\cite{damonte2019structural} &24.40 &23.60 & - &24.54 & 24.07 & - \\
			DCGCN~\cite{guo-etal-2019-densely} &25.9 &- &- &27.9 &- &57.3   \\
			G-Trans-F~\cite{zhu2019modeling} &27.23 &34.53 &61.55 &30.18 & 35.83 &63.20  \\
			G-Trans-SA~\cite{zhu2019modeling} &29.66 &35.45 &63.00 &31.54 & 36.02 &63.84 \\
			G-Trans-C~\cite{cai2020graph} &27.4 &32.9 &56.4 &29.8 &35.1 &59.4 \\
			G-Trans-W~\cite{wang2020amr} &25.9 &- &- &29.3 &- &59.0 \\
			\midrule
			G-Trans-F-Ours &30.15 &35.36 &63.08 &31.93  &37.23  &64.20 \\
			\quad + Node Prediction &30.72 & 35.94 &63.56  &32.99  &37.33 &64.53 \\
			\quad + Edge Prediction &31.07 & 35.87 &63.73  &33.44  &37.45 &64.62	\\
			\quad + Both Prediction  &\textbf{31.48} &\textbf{36.15} &\textbf{63.87} &\textbf{34.19}  &\textbf{38.18} &\textbf{65.72} \\
			\bottomrule
		\end{tabular}
		\caption{Main test results on LDC2015E86 and LDC2017T10.}
		\label{tab:allres}
	}
\end{table*}
Table~\ref{tab:allparam} lists all model hyperparameters used for experiments.
Specifically, we share the vocabulary of AMR node BPEs and target word BPEs.
Our implementation is based on the model of \citet{zhu2019modeling}, which is available at \url{https://github.com/Amazing-J/structural-transformer}.
Our re-implementation and the proposed model are released at \url{https://github.com/muyeby/AMR-Backparsing}.

\subsection{More Results}
We compare our model with more baselines and use more evaluation metrics (BLEU~\cite{papineni-2002-bleu}, Meteor~\cite{banerjee-lavie-2005-meteor,denkowski-lavie-2014-meteor} and CHRF++~\cite{popovic-2017-chrf}).
The results are shown in Table~\ref{tab:allres}.
It can be observed that our approach achieves the best performance on both datasets regardless of the evaluation metrics. 
This observation is consistent with Table~\ref{tab:main}.
\end{document}